\UseRawInputEncoding
\documentclass{article}

\usepackage[english]{babel}

\usepackage[letterpaper,top=2cm,bottom=2cm,left=3cm,right=3cm,marginparwidth=1.75cm]{geometry}

\usepackage{amsmath}
\usepackage{graphicx}
\usepackage[colorlinks=true, allcolors=blue]{hyperref}

\title{Indoor Air Quality Detection Robot Model Based on the Internet of Things (IoT)}

\author{Anggiat Mora Simamora, Asep Denih, Mohamad Iqbal Suriansyah}

\begin{document}
\maketitle

\begin{abstract}
This paper presents the design, implementation, and evaluation of an IoT-based robotic system for mapping and monitoring indoor air quality. The primary objective was to develop a mobile robot capable of autonomously mapping a closed environment, detecting concentrations of CO$_2$, volatile organic compounds (VOCs), smoke, temperature, and humidity, and transmitting real-time data to a web interface. The system integrates a set of sensors (SGP30, MQ-2, DHT11, VL53L0X, MPU6050) with an ESP32 microcontroller. It employs a mapping algorithm for spatial data acquisition and utilizes a Mamdani fuzzy logic system for air quality classification. Empirical tests in a model room demonstrated average localization errors below 5\%, actuator motion errors under 2\%, and sensor measurement errors within 12\% across all modalities. The contributions of this work include: (1) a low-cost, integrated IoT robotic platform for simultaneous mapping and air quality detection; (2) a web-based user interface for real-time visualization and control; and (3) validation of system accuracy under laboratory conditions.
\end{abstract}

\section{Introduction}

Air plays a crucial role in supporting life and is generally classified into outdoor and indoor air \cite{prabowo2017}. Indoor air quality (IAQ) is vital for human health, considering that nearly 90\% of human activities occur indoors \cite{bivolarova2016}. In developing countries, approximately 400 to 500 million people are adversely affected by indoor air pollution \cite{depkes2011}. Key pollutants include Volatile Organic Compounds (VOCs), carbon dioxide (CO$_2$), and cigarette smoke \cite{prabowo2017}.

According to the National Institute for Occupational Safety and Health (NIOSH), major sources of indoor air pollution consist of inadequate ventilation (52\%), indoor contamination (16\%), outdoor contamination (10\%), microbes (5\%), building materials (4\%), and other factors (13\%) \cite{osha2017}. Poor ventilation leads to variations in air distribution and pressure, which allows hazardous gases to spread unevenly within a room \cite{widyantara2018}. Because air moves due to pressure differences (advection) and gas particles migrate from higher to lower concentrations (diffusion), single-point air quality measurements may yield less accurate results \cite{widyantara2018}.

\section{Methods}
\begin{figure}[ht]
  \centering
  \includegraphics[width=0.5\textwidth]{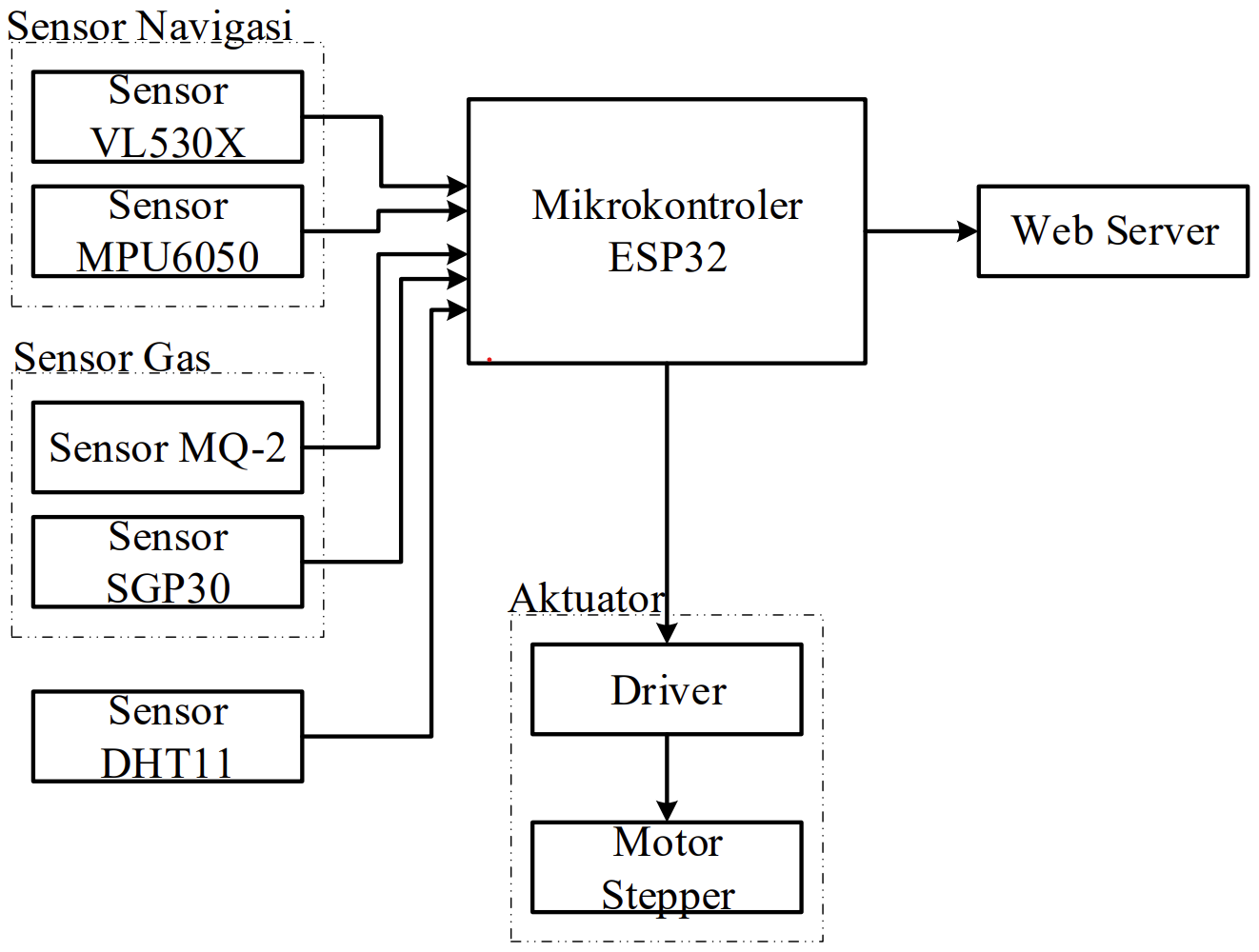} 
  \caption{System Block Diagram}
  \label{fig:diagram-blok}
\end{figure}

The proposed system consists of an ESP32-based mobile robot equipped with several sensors: SGP30 and MQ-2 for gas detection, DHT11 for temperature and humidity, MPU6050 for orientation, and VL53L0X for distance measurement. The robot is powered by a 3S 1600mAh LiPo battery and uses stepper motors controlled by A4988 drivers. A system block diagram is shown in Figure~ \ref{fig:diagram-blok}.

\subsection{Hardware and Software Design}

The robot’s hardware and mechanical parts were designed using SolidWorks and Kicad, with custom 3D-printed components and a PCB mainboard for wiring efficiency. The software was developed using Visual Studio Code, with the robot’s firmware running FreeRTOS for real-time multitasking. The web-based monitoring interface was built in JavaScript using P5.js to visualize data from the robot.

\begin{figure}[ht]
  \centering
  \includegraphics[width=1\textwidth]{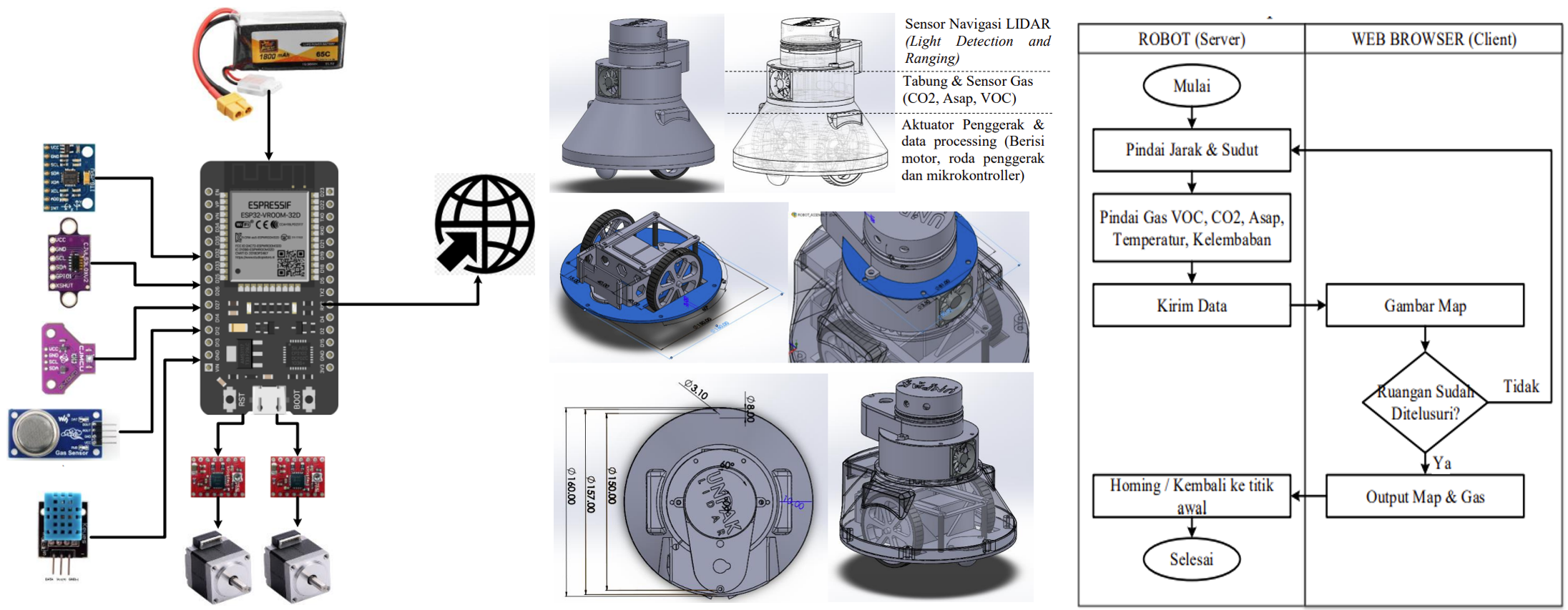} 
  \caption{Electronic, Mechanical \& Software Design}
  \label{fig:design-overview}
\end{figure}

\subsection{System Workflow}

The robot collects environmental data as it navigates a room, sending sensor readings and mapping data to the web interface via Wi-Fi in JSON format.

\subsection{Data Processing Methods}

Air quality classification (Good, Moderate, Poor) is performed using Mamdani fuzzy logic. Input variables include VOC, CO$_2$, smoke, temperature, and humidity, with rules and membership functions defined according to environmental standards. The centroid method is used for defuzzification.

In addition to air quality classification, spatial data from the robot’s mapping process is utilized to detect the geometry of the room. The data representing room walls displayed on the web user interface initially consists of discrete points. To extract meaningful wall structures, this study employs simple linear regression as the primary computational method. However, prior to regression analysis, the data must be sorted and grouped based on their spatial distribution along vertical and horizontal axes.

\subsubsection{Data Sorting Process}

\begin{figure}[ht]
  \centering
  \includegraphics[width=0.9\textwidth]{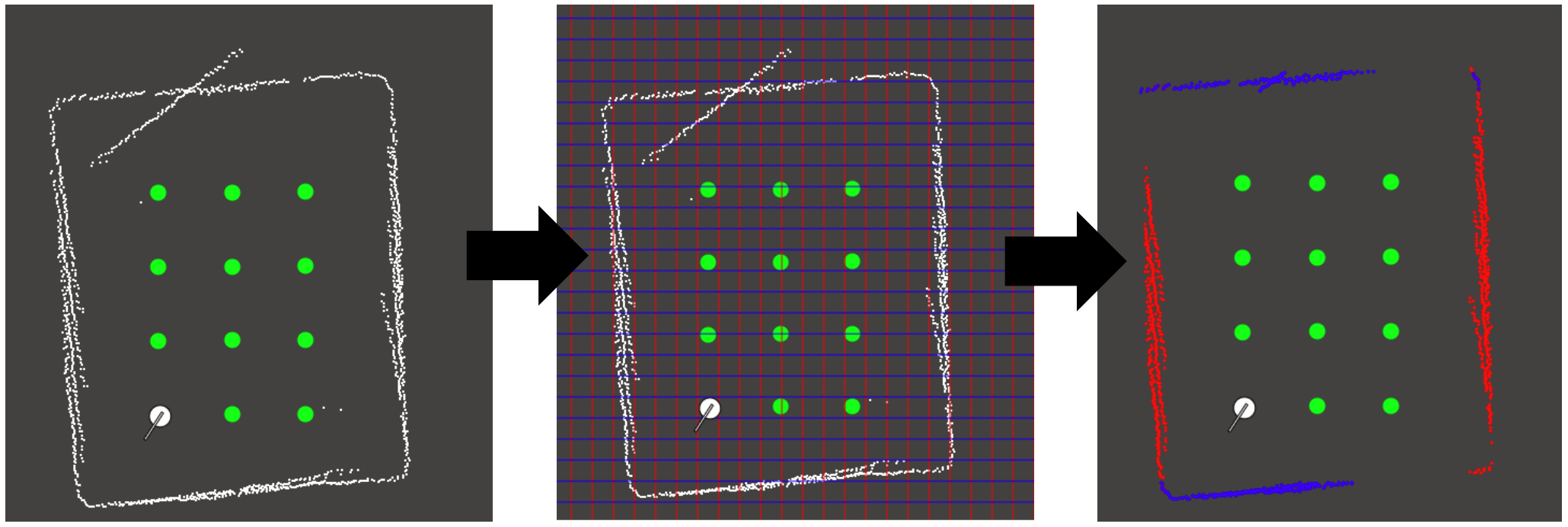}
  \caption{Data Sorting Process Based on Vertical and Horizontal Axes}
  \label{fig:data_sorting_process}
\end{figure}

In this step, the mapping data points are separated into two groups according to their orientation: vertical and horizontal clusters. This grouping facilitates more accurate regression calculations by isolating data points that are likely to belong to the same wall segment (see Figure~\ref{fig:data_sorting_process}).

\subsubsection{Simple Linear Regression Calculation}

After grouping, simple linear regression is applied to each data cluster to derive the linear function that best fits the points in that group. The regression model is defined as:

\[
Y = a + bX
\]

where:  
- $Y$ is the dependent variable,  
- $X$ is the independent variable,  
- $a$ is the intercept (constant),  
- $b$ is the slope (coefficient of $X$).

\begin{figure}[ht]
  \centering
  \includegraphics[width=0.65\textwidth]{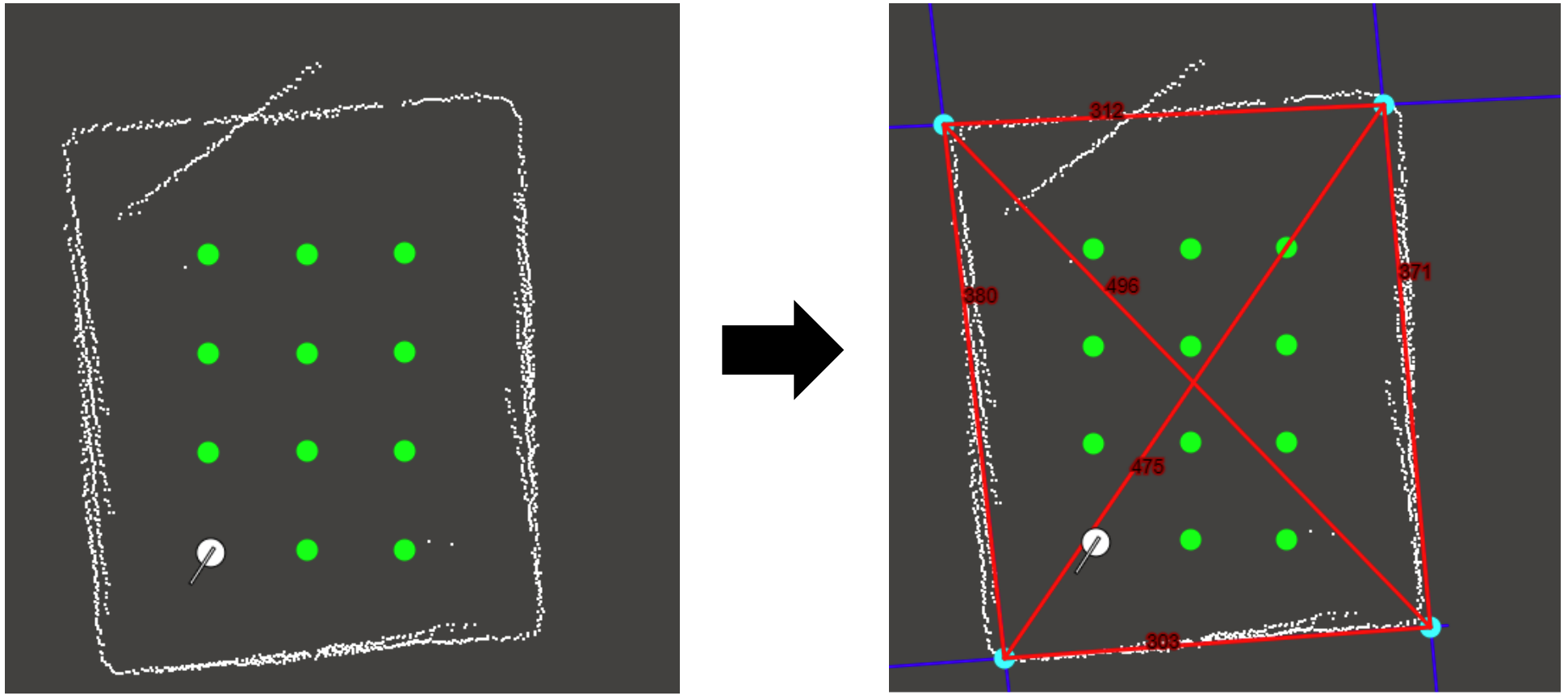}
  \caption{Simple Linear Regression Applied to Grouped Mapping Data}
  \label{fig:linear_regression_process}
\end{figure}

This computation yields several straight lines representing probable walls within the room. These regression lines are visualized as blue lines in Figure~\ref{fig:linear_regression_process}.

The intersections of these regression lines correspond to the corners of the room, marked by light blue points in the figure.

\subsubsection{Evaluation Process}

The output of this regression-based wall detection closely resembles the raw mapping data, indicating that the method effectively captures the room’s geometry. Consequently, the calculated lengths of walls and corner positions can be reliably used for further spatial analysis and navigation purposes.

\section{Results and Discussion}

Each sensor integrated into the robot was validated by comparing its readings against reference measurement devices. Specifically, the VOC sensor was validated using a CCS811, CO$_2$ with an MQ-135, smoke with an optical GP2Y1010AU0F, temperature and humidity with an AHT10, and battery voltage with a multimeter. The average error rates for these sensors were: VOC 10.95\%, CO$_2$ 10.63\%, smoke 11.68\%, temperature 9.61\%, humidity 4.46\%, and battery voltage 2.44\%. The distance sensor, validated using manual measurement tools, exhibited an average error of 20.06\%. Overall, these sensors provided sufficiently accurate readings to support effective indoor air quality monitoring. Table~\ref{tab:sensor_validation} summarizes the sensor validation error rates.

In addition to sensor validation, the robot’s mapping accuracy was evaluated by comparing its reported coordinates, wall dimension estimations, and homing capabilities against actual physical measurements. The average errors observed were 4.17\% and 4.23\% for gas coordinate points along the X and Y axes respectively, 5.39\% for wall dimension mapping, and a homing distance error of 9.09 cm. These results indicate reliable autonomous navigation within a controlled indoor environment. Table~\ref{tab:mapping_navigation} details the mapping and navigation accuracy metrics.

To assess the robustness of the Mamdani fuzzy logic method in compensating sensor inaccuracies, air quality classification results before and after applying fuzzy logic were compared. The average classification error decreased significantly from 9.47\% to 1.92\%, demonstrating the method’s effectiveness in improving reliability despite minor sensor errors.

The integration of multi-sensor data fusion, real-time spatial mapping, and fuzzy logic classification allows the robot to perform reliable indoor air quality monitoring alongside accurate room mapping. Throughout testing, all principal functionalities—including autonomous navigation, sensor integration, and web-based visualization—performed as intended.

\begin{table}[ht]
\centering
\caption{Sensor Validation Error Rates}
\label{tab:sensor_validation}
\begin{tabular}{lcc}
\hline
\textbf{Sensor}        & \textbf{Reference Device}       & \textbf{Average Error (\%)} \\
\hline
VOC                    & CCS811                        & 10.95                       \\
CO$_2$                 & MQ-135                        & 10.63                       \\
Smoke                  & Optical GP2Y1010AU0F          & 11.68                       \\
Temperature            & AHT10                         & 9.61                        \\
Humidity               & AHT10                         & 4.46                        \\
Battery Voltage        & Multimeter                    & 2.44                        \\
Distance Sensor        & Manual measurement            & 20.06                       \\
\hline
\end{tabular}
\end{table}

\vspace{1cm}

\begin{table}[ht]
\centering
\caption{Mapping and Navigation Accuracy}
\label{tab:mapping_navigation}
\begin{tabular}{lc}
\hline
\textbf{Parameter}               & \textbf{Average Error}    \\
\hline
Gas Coordinates (X-axis)         & 4.17\%                   \\
Gas Coordinates (Y-axis)         & 4.23\%                   \\
Wall Dimension Mapping           & 5.39\%                   \\
Homing Distance Error            & 9.09 cm                  \\
\hline
\end{tabular}
\end{table}

\section{Conclusion}

This study developed an indoor air quality detection robot that demonstrated satisfactory performance, with results successfully visualized on a web-based user interface. However, the air quality measurements at each point showed an average error of 5.56\% compared to actual values. The measurement errors were primarily influenced by sensor inaccuracies: VOC sensor exhibited an error of 10.95\%, CO$_2$ sensor 10.63\%, smoke sensor 11.68\%, temperature sensor 9.61\%, and humidity sensor 4.46\%. 

The impact of each sensor's error on the Mamdani fuzzy logic processing was quantified as follows: VOC 0.21\%, CO$_2$ 1.78\%, smoke 2.25\%, temperature 1.49\%, and humidity 3.84\%. Regarding mapping accuracy, the average errors were 4.17\% for the X-coordinate of gas location, 4.23\% for the Y-coordinate, and 5.39\% for wall dimension estimations. These discrepancies were influenced by the robot's movement resolution (1 mm per step) and rotational sensitivity (1° per turn), as well as limitations in the distance and orientation sensors, which occasionally caused deviations between the actual environment and the mapped data.

The mapping functionality relies on a continuous connection between the robot and the web UI; loss of connection results in halted robot operation due to interrupted command reception. Sensor data are encoded in JSON format and transmitted to the web UI for real-time processing, logging, and visualization as an interactive map. Users can also download the data as JSON files for further analysis.

The source code and algorithms developed in this study are publicly available at the GitHub repository: \url{https://github.com/anggiatm/ROBOT_GAS_MAPPING}. This resource may serve as a valuable reference for future research and development in robotic air quality monitoring systems.

\bibliographystyle{alpha}
\bibliography{sample}

\end{document}